\documentclass[journal]{IEEEtran}
\ifCLASSINFOpdf
\else
\fi
\usepackage[utf8]{inputenc}
\usepackage[T1]{fontenc}
\usepackage{graphicx}
\usepackage{booktabs}
\usepackage{adjustbox}
\usepackage{multirow}
\usepackage[caption=false]{subfig}
\usepackage{array}
\usepackage[ruled,vlined,linesnumbered]{algorithm2e}
\usepackage{amsmath}
\usepackage{url}
\DeclareMathOperator{\minC}{minC}

\SetKw{Break}{break}

\begin{document}
%
\title{Automated Design of Heuristics for the Container Relocation Problem}
%
%
%

\author{Marko~Đurasević,~\IEEEmembership{Member,~IEEE,}
        and~Mateja~Đumić
\thanks{Marko Đurasević is with the Department of Electronics, Microelectronics, Computer and Intelligent Systems, Faculty of Electrical Engineering, University of Zagreb, Zagreb 10000, Croatia, e-mail: marko.durasevic@fer.hr.}
\thanks{Mateja Đumić is with the Department of Mathematics, J. J. Strossmayer University of Osijek, Osijek 31000, Croatia, e-mail: mdjumic@mathos.hr.}
}
 
%
%

\markboth{Transactions on Evolutionary Computation}%
{Shell \MakeLowercase{\textit{et al.}}: Bare Demo of IEEEtran.cls for IEEE Journals}
%


\maketitle

\begin{abstract}
The container relocation problem is a challenging combinatorial optimisation problem tasked with finding a sequence of container relocations required to retrieve all containers by a given order. Due to the complexity of this problem, heuristic methods are often applied to obtain acceptable solutions in a small amount of time. These include relocation rules (RRs) that determine the relocation moves that need to be performed to efficiently retrieve the next container based on certain yard properties. Such rules are often designed manually by domain experts, which is a time-consuming and challenging task. This paper investigates the application of genetic programming (GP) to design effective RRs automatically. The experimental results show that GP evolved RRs outperform several existing manually designed RRs. Additional analyses of the proposed approach demonstrate that the evolved rules generalise well across a wide range of unseen problems and that their performance can be further enhanced. Therefore, the proposed method presents a viable alternative to existing manually designed RRs and opens a new research direction in the area of container relocation problems.
\end{abstract}


%
\IEEEpeerreviewmaketitle

\section{Introduction}
%
%
%
%
\IEEEPARstart{T}{he} container relocation problem (CRP) is an important combinatorial optimisation problem that appears in warehouse and yard management. Nowadays, this problem is gaining more importance since most international trade is carried out by the international shipping industry~\cite{Jovanovic2019}.

Usually, containers are placed in a stacking area while waiting for loading. Because of the limited capacity of the stacking area, containers are placed side by side or on top of each other. By storing containers in such a way, blocks are formed. Each block consists of a number of stacks (width), a number of tiers (height), and a number of bays (length).

Loading of containers is done in a predetermined order. If the container that needs to be retrieved next is not on the top of its stack, all containers above it need to be relocated. The relocation of containers is, in most cases, inevitable because of the incomplete or unavailable information. In \cite{Steenken2004} authors give an estimation that 30-40\% of the outbound containers at European terminals do not have correct information about the ship or the destination port. The situation is even worse for inbound containers, and only 10-15\% of container shipment is known in advance \cite{Steenken2004}. 
 

CRP deals with the relocation and retrieval of containers at the same time. In literature, a variety of approaches used for solving different variants of CRP can be found. CRP variants differ in restrictions on moves, retrieval priorities of blocks (whether they are distinct or not), whether all blocks need to be retrieved, how many blocks can be moved at a time, and whether the stacks are ordered in a single or multiple bays. A formal problem classification can be found in \cite{Lu2020}. 

The single bay CRP was for the first time introduced by Sculli and Hui in 1988 \cite{SCULLI1988585}. Avriel et al. \cite{Avriel2000} showed that it is an NP-complete problem. In \cite{Kim2006} a heuristic based on the expected number of additional relocations (ENAR), which was used in a branch-and-bound algorithm to search for the optimal solution, was introduced. A beam search algorithm and three relocation rules (RRs): lowest position (TLP), reshuffle index (RI), and reshuffle index with look-ahead (RIL) were proposed in \cite{Wu2010}. In \cite{Diaz2017} a GRASP-based algorithm that found new bounds for many problem instances was proposed. Lee and Lee \cite{Lee2010} developed a three-phase heuristic for multibay CRP, which was the first approach in the literature applied for multiple bays. In this problem, besides the number of relocations, it is essential to reduce the crane working time, which can be significant when moving from one bay to another.
 
A corridor method combined with dynamic programming was used in \cite{Caserta2011}, while in \cite{Caserta2012}, two mathematical models and the Min-Max rule were introduced. In \cite{Forster2012} a tree search procedure for multibay CRP was introduced. This approach defines well placed and bad placed containers, which can help calculate a lower bound and choose moves. Iterative Deepening A* algorithm was used in \cite{Zhu2012} for solving CRP. In this paper, three lower bounds and four heuristics were used in the nodes of a tree. A new heuristics which, besides the container that needs to move, takes into account the properties of the container that needs to move next is introduced in \cite{Jovanovic2014}. In \cite{LIN2015132} authors propose the heuristic based on groups to solve CRP with multiple bays. In this paper, cranes that can move one or more containers at a time were observed.

Authors in \cite{Borjian2015} used the A* algorithm for solving CRP exactly and approximately. They also study the average-case asymptotic behaviour of the CRP when the number of stacks increases. CRP with time window was studied in \cite{KU20161031} in which abstraction heuristic is developed to improve the tree-search approach used for solving it. In most recent research, four rules combined with a genetic algorithm were used to determine the best sequence of container retrievals \cite{Maglic2019}, and GRASP for multibay CRP is introduced \cite{Cifuentes2020}. Also, the dynamic stacking problem with uncertainties was studied in \cite{Raggl2020}. In this paper, the problem was solved using two approaches: hand-crafted rules and model-based.  The second one solves the problem by solving a static model of a specific planning horizon and seems to be better of these two approaches.

From the previous overview, it is evident that in most research, heuristic methods have been applied for solving the CRP, which ranged from simple heuristics to more complex metaheuristics. Metaheuristics have the benefit that they achieve better results than simple heuristic rules at the expense of a longer execution time, especially as the problem size increases. On the other hand, simple RRs obtain good solutions in an almost negligible time since they do not traverse the search space in the quest for good solutions but rather construct a good solution iteratively using a certain strategy. These strategies are manually defined and include domain knowledge defined by experts. However, effective RRs are challenging to design and require good expert knowledge about the problem.

This study investigates the automatic construction of RRs using genetic programming (GP) to mitigate the previous problem. GP is an evolutionary computation method commonly used for automatic heuristic generation \cite{Burke2013}, which already demonstrated good performance for various scheduling problems \cite{Branke2016}. In this study, GP is applied for constructing simple RRs, which iteratively determine how the containers should be relocated. The experimental analysis demonstrates that automatically generated RRs achieve a better performance than existing rules on an extensive benchmark dataset. Furthermore, a detailed analysis of different aspects of the proposed approach is performed to understand them better. 

The remaining part of this paper is organised as follows. In Section II, the problem description is given. Section III describes the design of RRs. The experimental setup is given in Section IV, while experimental results are presented in Section V. Section VI provides further analyses. Finally, Section VII concludes the paper and outlines future research directions.

\section{Problem description}

In this paper, the single bay CRP will be considered. The bay consists of $S$ stacks with $T$ tiers. The height of each stack $s$ is denoted with $h(s)$, and it has to be smaller or equal to $T$. The assumptions are that all containers are of the same size, the gantry crane can move only one container at a time, all containers must be retrieved, and each container has a unique priority. 

\begin{figure}[]
\centering
\includegraphics[width=8cm]{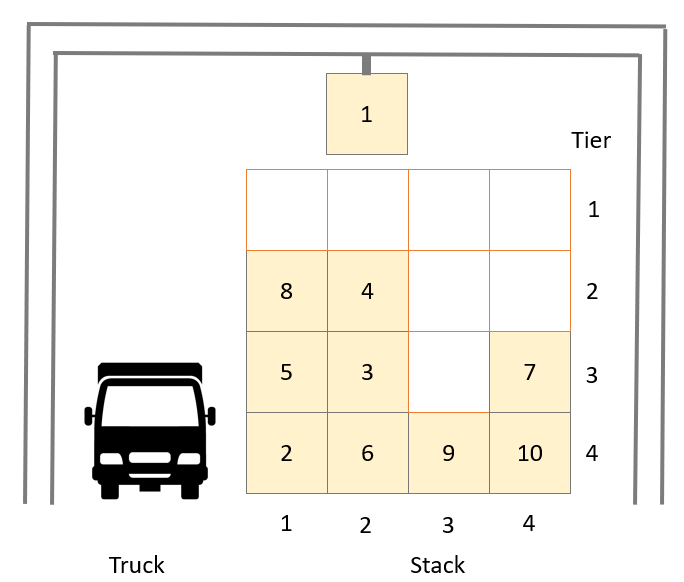}
\caption{Example of a container bay}
\label{fig:CRP_example}
\end{figure}

As previously mentioned, CRP consist of two types of operations: relocation and retrieval. Relocation is the operation of moving the container from the top of one stack to another one. The container can be relocated to a stack only if the stack height is smaller than $T$. Retrieval is the operation of picking a container from the top of the stack and moving it on the truck used for loading. The truck to which the container is retrieved is expected to be located at the beginning of the bay, at position 0. The gantry crane, which is located in the bay, performs the relocation and retrieval operations. An example of container bay can be seen in Figure~\ref{fig:CRP_example}.

Each container is given an ID that determines the order in which they need to be retrieved. The container with the smallest ID needs to be retrieved first and is called the \emph{target container}. If the target container is not on the top of the stack, relocations needs to be performed. The stack from which the container is moved is called the \emph{origin stack}, while the stack to which it is moved is called the \emph{destination stack}. 

In the example given with Figure~\ref{fig:CRP_example}, the container with ID 1 was on the top of the stack, and it is in the process of retrieval, while the container with ID 2, which needs to be retrieved next, is blocked. Containers with IDs 8 and 5 need to be relocated to retrieve the container with ID 2. If one of these containers is relocated to stack two, it will become full, and further containers cannot be relocated to it.

Feasible solutions of the CRP are all sequences of relocations and retrievals that ensure that the crane can retrieve all containers in a predetermined order. The goal is to find the one that minimises the given objective. Usually, the number of relocations or total crane operation time are used as objectives.

When using the crane operation time as an objective, it is important to consider all parts of relocating and retrieving containers. For each movement, the time needed for the crane trolley to come to the container, pick it up and move to the destination stack needs to be calculated. The speed of moving crane trolley per container width (moving to the adjacent stack) is $1.2$ s, while pick-up time is $30$ s \cite{Cifuentes2020}.  The time invested in placing the crane to the origin stack and moving the container from the origin to the destination stack will be $1.2\times (number\,\, of\,\, passed\,\, stacks)$. Consequently, the time for relocating the container can be calculated as: $1.2\times|crane\,\,location - origin \,\,stack| + 1.2\times|origin \,\,stack - destination \,\,stack | + pickup \,\,time $. The total crane operation time is the sum of times needed for each movement.

In the literature, two types of CRP are distinguished: the restricted CRP and the unrestricted CRP. In the restricted CRP, only containers that are above the target container can be relocated, while in the unrestricted CRP, it is possible to relocate containers from all stacks. In this study, both variants are considered. Using the classification of \cite{Lu2020}, the problems considered in this study can be denoted as res|dis|com|ind and unr|dis|com|ind, which means that either restricted or unrestricted moves are considered, with distinct priorities, complete retrieval, and individual container moves. 

\section{Design of relocation rules}
RRs can be divided into two components: a relocation scheme (RS) that defines the overall strategy of how containers will be moved and which stacks are eligible for selection, and a priority function (PF) used to rank the available stacks. The RS uses the PF to rank all the stacks and selects the best one to current container will be relocated. Since the RS has to consider some constraints and is usually straightforward, it is defined manually. However, the PF which performs the ranking can be complex, and therefore it is generated using GP. 

The RS can be defined in different ways. However, all schemes function similarly. If the target container is on the top of its stack, it can immediately be removed from the yard. Otherwise, if other containers are on top of the target container, they need to be relocated to other stacks so that the target container is on top. Several relocation schemes are proposed to test how different decisions can affect the performance of RRs. However, all of them can be categorised as restricted or unrestricted versions, depending on which kind of relocation moves are allowed. 

Algorithm~\ref{alg:restricted} outlines the restricted RS variant, which will be denoted as RE. In this scheme, the origin stack is the stack where the target container is located, whereas the destination is any other stack that is not full. For each other stack, the priority is calculated using the PF, and the container on top of the origin stack is relocated to the stack with the lowest priority value. After a certain number of relocations, the target container will be on top and can be retrieved. The entire procedure is repeated until the yard is empty. Based on this scheme, an alternative one denoted as REN will also be used. When calculating the priorities for the stack, this scheme skips the stack with the next target container to prevent a possible relocation of a container to that stack. 

\begin{algorithm}[]
\SetAlgoLined 
\While{Container yard not empty}{
 C = GetTargetContainer()\;
 S = GetStack(C)\;
 \While{container C is not on top of stack S}{
  \ForEach{stack st, st != S and st not full}
  {
    $\pi_s$ = CalculatePriority(S);
  }
  Relocate top container from S to the stack with $min(\pi_s)$\;
 }
 Retrieve container C\;
 }
 \caption{Restricted relocation scheme}
 \label{alg:restricted}
\end{algorithm}

Algorithm~\ref{alg:unrestricted} shows the basic unrestricted scheme denoted as UN. The $\minC$ function returns the lowest container ID on a stack. The main difference between the UN and RE schemes is that after the destination stack is determined, it is checked whether the stack contains containers with an ID that is smaller than the ID of the container that will be relocated to that stack. If such a container exists, the top container will be relocated again at some point in the future. Therefore, this scheme tries to relocate these containers from the destination stack to stacks where all containers have a larger ID, if they exist. When all these containers have been relocated, or no such stack exists, the procedure proceeds as the RE scheme. 

\begin{algorithm}[]
\SetAlgoLined
\While{Containers remain}{
 C = GetNextContainer()\;
 S = GetStack(C)\;
 \While{container C is not on top of stack S}{
  \ForEach{stack st except S}
  {
    $\pi_s$ = CalculatePriority(S);
  }
  D = stack with $min(\pi_s)$\;
  SC = top container of stack S\;
  DC = top container of stack D\;
  \While{true}{
    \eIf{Stack S1 exists such that $\minC(S1)> DC$}{
        Relocate container DC to S1\;
   }{
    \Break \;
  }
  }
  Relocate top container from S to the stack with $min(\pi_s)$\;
 }
 Retrieve container C\;
 }
 \caption{Unrestricted relocation scheme}
 \label{alg:unrestricted}
\end{algorithm}
 
Except for this basic unrestricted scheme, three alternative schemes were also defined. The UNC variant works similarly as the RE scheme; however, instead of only calculating the priority for the destination stack, it calculates the priority for every two pairs of stacks, in which the first one represents the origin stack and the second one the destination stack. The pair with the lowest priority value is selected, and the top stack from the origin stack is relocated to the destination stack. Since it is possible that such a strategy could lead to infinite relocations, this procedure is done a certain number of times in each iteration, after which the scheme acts as the UN scheme in the sense that it only relocates containers from the stack where the target container is located. The UNC2 scheme functions similarly, except that it used two expressions, one to select the destination container and the other one to select the origin container of the relocation. Finally, the UNP scheme works in the same way as the UN; however, it evolves an additional expression used to determine the destination of the unrestricted moves, i.e., to which stacks the containers from the destination stack will be relocated. 
 
As previously outlined, the priority function is evolved by GP. For that purpose, the primitive set needs to be specified. The function set consists of summation, subtraction, multiplication, and protected division (returns 1 if the divisor is close to 0). Although other functions were tested (minimum, maximum, negation, and if), their inclusion did not lead to any improvements in the results, and as such, were not used. The terminal nodes that were used are denoted in Table~\ref{tbl:terminals}. The proposed set of terminals includes different simple properties (stack height, number of empty positions, ID of the current container) but also more complicated nodes (number of containers with a lower ID than the current container, average of container IDs on stack).

\begin{table}[]
\centering
\caption{List of the applied terminal nodes}
\label{tbl:terminals}
\begin{tabular}{@{}lp{7.5cm}@{}}
\toprule
Name  & Description                                                                                                       \\ \midrule
SH    & Stack height                                                                                                      \\
EMP   & Number of empty places in stack                                                                                   \\
CUR   & ID of the current container that will be relocated                                                                \\
DUR   & Time required to transfer the container to the desired stack                                                      \\
RI    & Number of containers with a smaller ID on the stack than the ID of selected container                                   \\
MIN   & The smallest container ID of the selected stack                                                                   \\
AVG   & Average ID of containers in a stack                                                                               \\
REM   & Remaining containers until the container which needs to be removed is on top                                      \\
NEXT  & Returns 1 if the selected stack contains the next container that will be retrieved and 0 otherwise                \\
DIFF  & Difference between the ID of the container with the lowest ID on the stack and the one that needs to be relocated \\
EMPTY & Returns 1 if a stack is empty and 0 otherwise                                                                     \\
WL    & Number of containers on the stack for which no container of a larger ID is on top of them                     \\
NL    & Number of containers on the stack for which a container of a larger ID is on top of them                      \\
DSM   & Height of the first container in a stack with a smaller ID than that of the considered container                                     \\ \bottomrule
\end{tabular}
\end{table}

\section{Experimental setup}

The Caserta~\cite{Caserta2011} and Zhu~\cite{Zhu2012} datasets are used to test the proposed method. In the Caserta dataset, the number of stacks and containers per stack can be between 3 and 10. All stacks have the same initial height in all problem instances. The maximum allowed height of a stack is $h+s$, where $h$ represents the stack height at the start. The Zhu dataset contains problems with 6 to 10 stacks and between 15 and 69 containers in the bay. The difference in comparison to the Caserta dataset is that the maximum stack height is specified for each instance in this dataset. Usually, the problem instances in the Zhu dataset have a smaller maximum height and consequently restrict the possible relocation moves at the start. The Caserta dataset contains 840 instances, whereas the Zhu dataset contains 125000 instances. 

In the experiments, three independent datasets were used, train, validation, and test. The train set was used to generate the expressions for the RRs, the validation for parameter tuning, and the test set to determine the final quality of the evolved rules. For the test set, the original Caserta and Zhu sets are used with all instances. The train and validation sets are generated randomly using the procedures described in~\cite{Caserta2011, Zhu2012}. For the Caserta dataset, those sets contained the same number of instances as the original (840 instances). On the other hand, for the Zhu instances, these sets contained 1000 problems as matching the size of the original set would significantly increase the training time. 

All GP parameters were optimised in preliminary experiments. After these experiments, the population size was set to 1000 individuals, the mutation probability to 0.3 for the restricted and 0.1 for the unrestricted schemes, the tree depth to 5, and the termination criterion to 50000 function evaluations. Regarding the genetic operators, for crossover the subtree, uniform, context preserving, size fair, and one point operators were used, whereas for mutation the subtree, hoist, node complement, node replacement, permutation, and shrink mutation were used. Since several genetic operators are defined, in each iteration, the algorithm randomly selects, with an equal probability, an operator with which it will perform the crossover and mutation of individuals. When GP had to evolve two PFs, the GP individual consisted of 2 expressions on which genetic operators were executed independently from each other. 

All experiments were executed 30 times, and the minimum, median, and maximum values of these executions are denoted in the results. Two optimisation criteria are considered, namely the number of relocations and the crane execution time. The Mann-Whitney test will be used to compare whether there exists a statistically significant difference. The non-parametric test was used since the normality of all data could not be confirmed by using the Shapiro-Wilk test.  

The proposed method will be compared to seven manually designed RRs from the literature. These include rules which perform only restricted moves, namely TLP \cite{Lee2010}, RI \cite{Lee2010}, Min-Max \cite{Caserta2012}, PR3 \cite{Zhu2012}, PR4 \cite{Zhu2012}, as well as rules that use unrestricted moves like PU1 and PU2 \cite{Zhu2012}.

The problem instances and the source code of the methods can be obtained from \url{http://www.zemris.fer.hr/~idurasevic/CRP/CRP.7z}.

\section{Results}

The results obtained by the proposed method for the Caserta dataset are shown in Table~\ref{tbl:rescaserta}. The first part of the table outlines the results obtained by the manually designed RRs. The second part of the table shows the results obtained by automatically designed RRs with different relocation schemes when optimising the two objectives. The values for both criteria are denoted for the obtained rules regardless of which criterion was optimised to gain a better insight into how rules evolved for one criterion perform on the other. The best results obtained in the table are denoted in bold.

The obtained results demonstrate that the automatically designed RRs in most cases outperform all the manually designed RRs. This is evident because, for all RSs except UNP, the worst rules outperform all manually designed RRs. This shows that the proposed method is stable and can easily evolve new rules that outperform existing manually designed rules. When considering only the restricted rules, the best-generated rule performs better by 7.7\% for the relocation criterion and 4.3\% for the crane operation time criterion than the best manually designed rule PR4. For the unrestricted variant, the best-generated rule achieved 5.1\% for the number of relocations and 3.8\% for the crane operation time compared to the best manually designed rule (PU2). One thing that can be observed from the results is that when optimising the number of container relocations, the values for the crane operation time are similar to the values obtained when optimising this objective. However, the reverse is not always true, which might mean that optimising the number of container relocations also implicitly optimises the crane operation time. 

\begin{table}[]
\centering
\caption{Results obtained on the Caserta dataset}
\label{tbl:rescaserta}
\begin{tabular}{@{}lccc|ccc@{}}
\toprule
                     & \multicolumn{3}{c}{Container relocations} & \multicolumn{3}{c}{Crane operation time}    \\ \midrule
TLP                  & \multicolumn{3}{c}{35982}       & \multicolumn{3}{c}{2430770} \\
RI                   & \multicolumn{3}{c}{29524}       & \multicolumn{3}{c}{2162170} \\
MM                   & \multicolumn{3}{c}{28996}       & \multicolumn{3}{c}{2173290} \\
PR3                  & \multicolumn{3}{c}{25859}       & \multicolumn{3}{c}{2064600} \\
PR4                  & \multicolumn{3}{c}{25787}       & \multicolumn{3}{c}{2063070} \\
PU1                  & \multicolumn{3}{c}{25049}       & \multicolumn{3}{c}{2034230} \\
PU2                  & \multicolumn{3}{c}{24962}       & \multicolumn{3}{c}{2031130} \\ \midrule
\multicolumn{1}{c}{} & \multicolumn{6}{c}{Optimising crane relocations}                    \\
                     & \multicolumn{3}{c}{Container relocations} & \multicolumn{3}{c}{Crane operation time}    \\\midrule
                     & Min       & Med      & Max      & Min     & Med     & Max     \\ \midrule
RE           & 23816     & 24122    & 24508    & 1978200 & 1993387   & 2004644   \\
REN &     23849      &    24154      &    24449      &    1978772     &   1991568      &   2001708  \\
UN  &  \textbf{23679}     &      \textbf{23967}     &     \textbf{24122}     &     1975008     &              1985987    &    1996640     \\
UNC &    23777     &     24236      &    24579      &    1979806      &       1995411  &         1996640         \\
UNT         &    24041       &    24312      &    24709      &    1986174     &   1997805      &      2016432   \\
UNP         &      24341     &     24876     &     25405     &         1997278&    2019489     &       2038289  \\\midrule
                     & \multicolumn{6}{c}{Optimising crane operation time}                           \\
                     & \multicolumn{3}{c}{Container relocations} & \multicolumn{3}{c}{Crane operation time}    \\ \midrule
                     & Min       & Med      & Max      & Min     & Med     & Max     \\ \midrule
RE           & 23717     & 24298    & 24609    & 1974102 & 1992614   & 2001919   \\
REN &     23965      &    24276      &    24608      &    1982433     &   1991391      &   2005142  \\
UN         &       23757    &      24053    &      24463    &      \textbf{1975159}   &    \textbf{1985758}    &    \textbf{1995911}     \\
UNC         &     23850      &     24291     &     24684     &     1977746    &    1991795     &  2005178       \\
UNT         &     23924      &        24292  &    24674      &        1981213 &  1993391       &      2009551   \\
UNP         &      24389     &     24969     &     26741     &     1998595    &    2020428     &  2060607       \\ \bottomrule
\end{tabular}

\end{table}

\begin{figure}[]
\centering
\includegraphics[width=6cm]{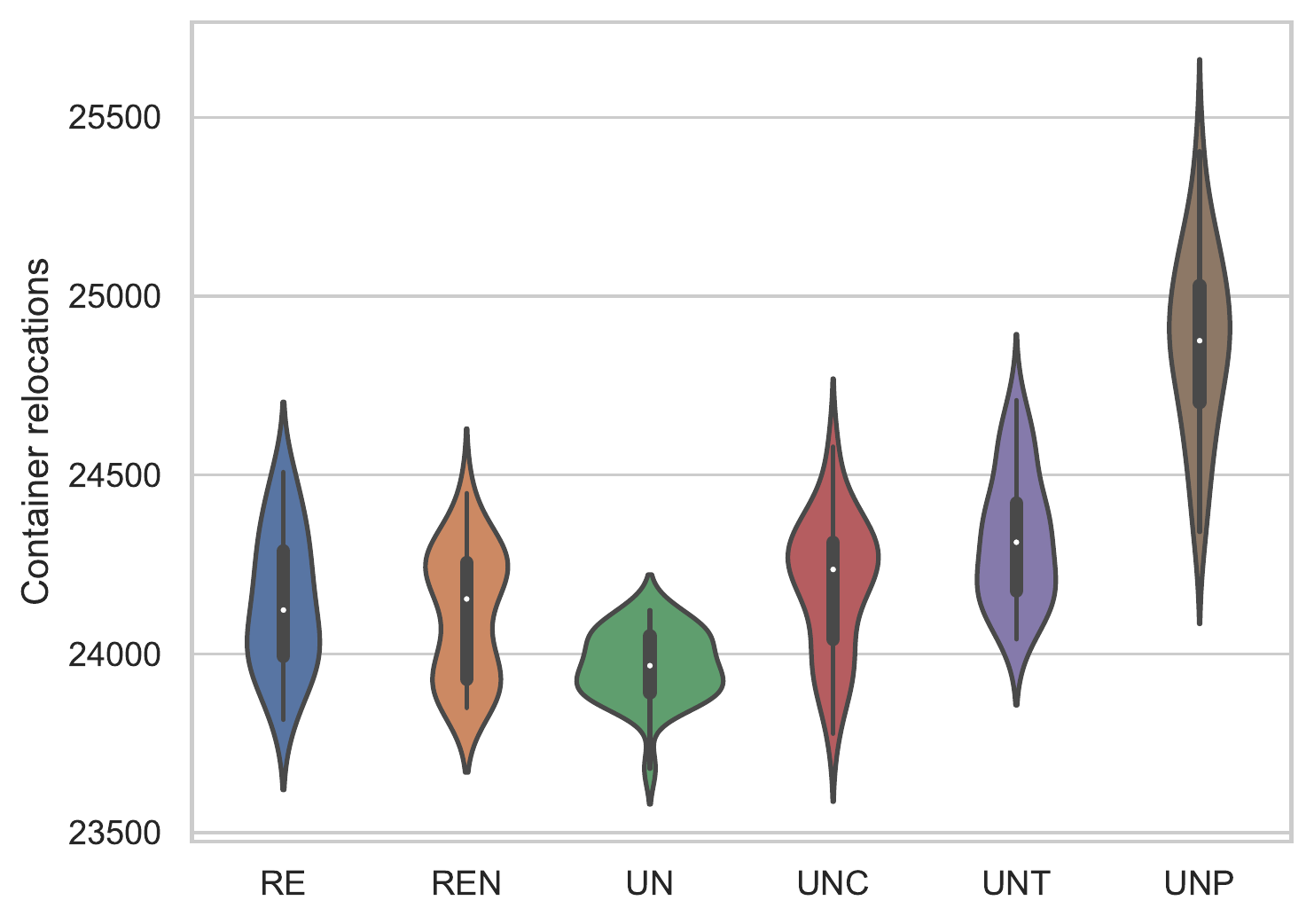}
\caption{Optimising the number of relocations on the Caserta dataset}
\label{fig:violincasrel}
\end{figure}

\begin{figure}[]
\centering
\includegraphics[width=6cm]{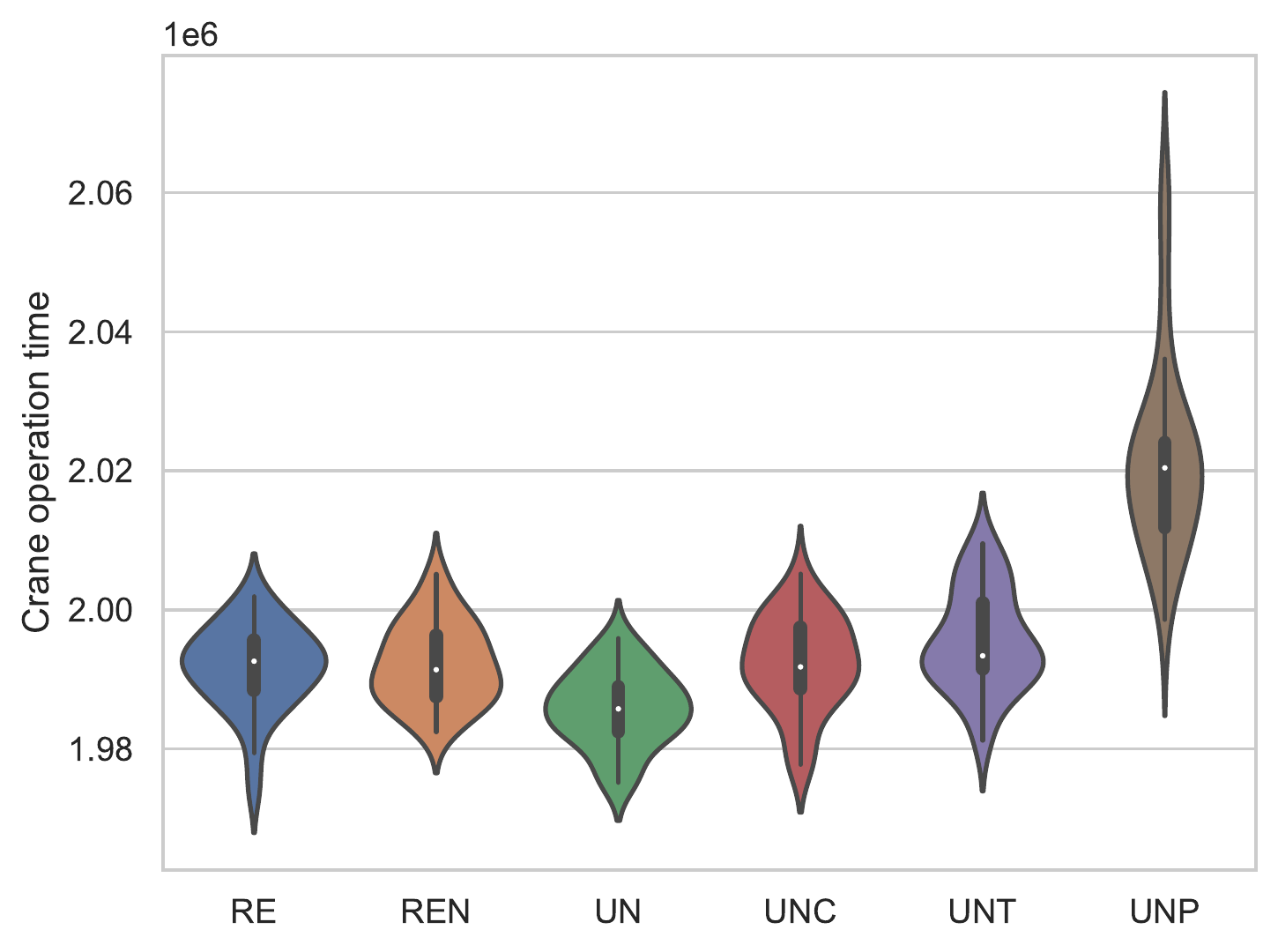}
\caption{Optimising the crane operation time on the Caserta dataset}
\label{fig:violincastime}
\end{figure}

To more easily compare the results, they are also illustrated using violin plots in Figures \ref{fig:violincasrel} and \ref{fig:violincastime}. These plots show that the UN scheme obtains the overall best results. When considering only the restricted schemes, it is clear that there is no difference between the RE and REN variants. On the other hand, the differences are much more evident for the unrestricted variants. In this case, the UN scheme obtains significantly better results than any other RS. 

Table \ref{tbl:zhu} shows the results obtained on the Zhu dataset. The results again demonstrate that most of the automatically generated RRs outperform the existing manually designed rules. In this case, the best-evolved rules for the restricted variant outperform the best existing rule (PR4) by 7\% for the number of relocations and 3.3\% for the crane operation time. For the unrestricted variant, the best manually designed rule achieves a better performance than the best manually designed rule (PU2) by 4\% for the number of relocations and 3\% for the crane operation time.

\begin{table}[]
\centering
\caption{Results obtained on the Zhu dataset}
\label{tbl:zhu}
\adjustbox{max width=\columnwidth}{%
\begin{tabular}{@{}lccc|ccc@{}}
\toprule
                     & \multicolumn{3}{c}{Container relocations} & \multicolumn{3}{c}{Crane operation time}    \\ \midrule
TLP                  & \multicolumn{3}{c}{551023}       & \multicolumn{3}{c}{41038200} \\
RI                   & \multicolumn{3}{c}{469502}       & \multicolumn{3}{c}{37508600} \\
MM                   & \multicolumn{3}{c}{473358}       & \multicolumn{3}{c}{38155300} \\
PR3                  & \multicolumn{3}{c}{436717}       & \multicolumn{3}{c}{37015300} \\
PR4                  & \multicolumn{3}{c}{435886}       & \multicolumn{3}{c}{36994500} \\
PU1                  & \multicolumn{3}{c}{423058}       & \multicolumn{3}{c}{36482700} \\
PU2                  & \multicolumn{3}{c}{422555}       & \multicolumn{3}{c}{36476100} \\ \midrule
\multicolumn{1}{c}{} & \multicolumn{6}{c}{Optimising container relocations}                    \\
                     & \multicolumn{3}{c}{Container relocations} & \multicolumn{3}{c}{Time}    \\\midrule
                     & Min       & Med      & Max      & Min     & Med     & Max     \\ \midrule
RE           & 413469     & 414885    & 417092    & 35920974 & 35989754 & 36228750   \\
REN           & 410993     & 414126    & 416897    & 35880669 & 35933913 & 36142543   \\
UN         &   \textbf{405571}        &     \textbf{407103}     &    \textbf{409266}      &   35625504      &     35681901    &       35926484  \\
UNC         &      413179     &    415247      &       417865   &      35897273   &    35982310     &      36267563   \\
UNT         &     412713      &       415548   &          419194&     35905778    &       36003943  &     36191417    \\
UNP         &      414597     &    418379      &       424787   &      35948660   &    36103360     &      36341047   \\
\midrule
                     & \multicolumn{6}{c}{Optimising crane operation time}                           \\
                     & \multicolumn{3}{c}{Container relocations} & \multicolumn{3}{c}{Crane operation time}    \\ \midrule
                     & Min       & Med      & Max      & Min     & Med     & Max     \\ \midrule
RE           & 414007     & 416447    & 422213    & 35815092 & 35907664 & 36067638   \\
REN           & 410645     & 415655    & 418710    & 35791829 & 35890043 & 36002463   \\
UN         &   406609        &     408286     &    412797      &   \textbf{3537753}     &  \textbf{35629511}      &    \textbf{35736631}     \\
UNC         &      414030     &    416100      &       428039   &      35821219   &    35915527     &  36106539       \\
UNT         &         414003  &   416458       &      639548    &    35797176    &       35956742  &     36399481    \\
UNP         &       415371    &         418549 &    424566      &   35900349      &         36030791&       35151525  \\ \bottomrule
\end{tabular}}
\end{table}

\begin{figure}[]
\centering
\includegraphics[width=6cm]{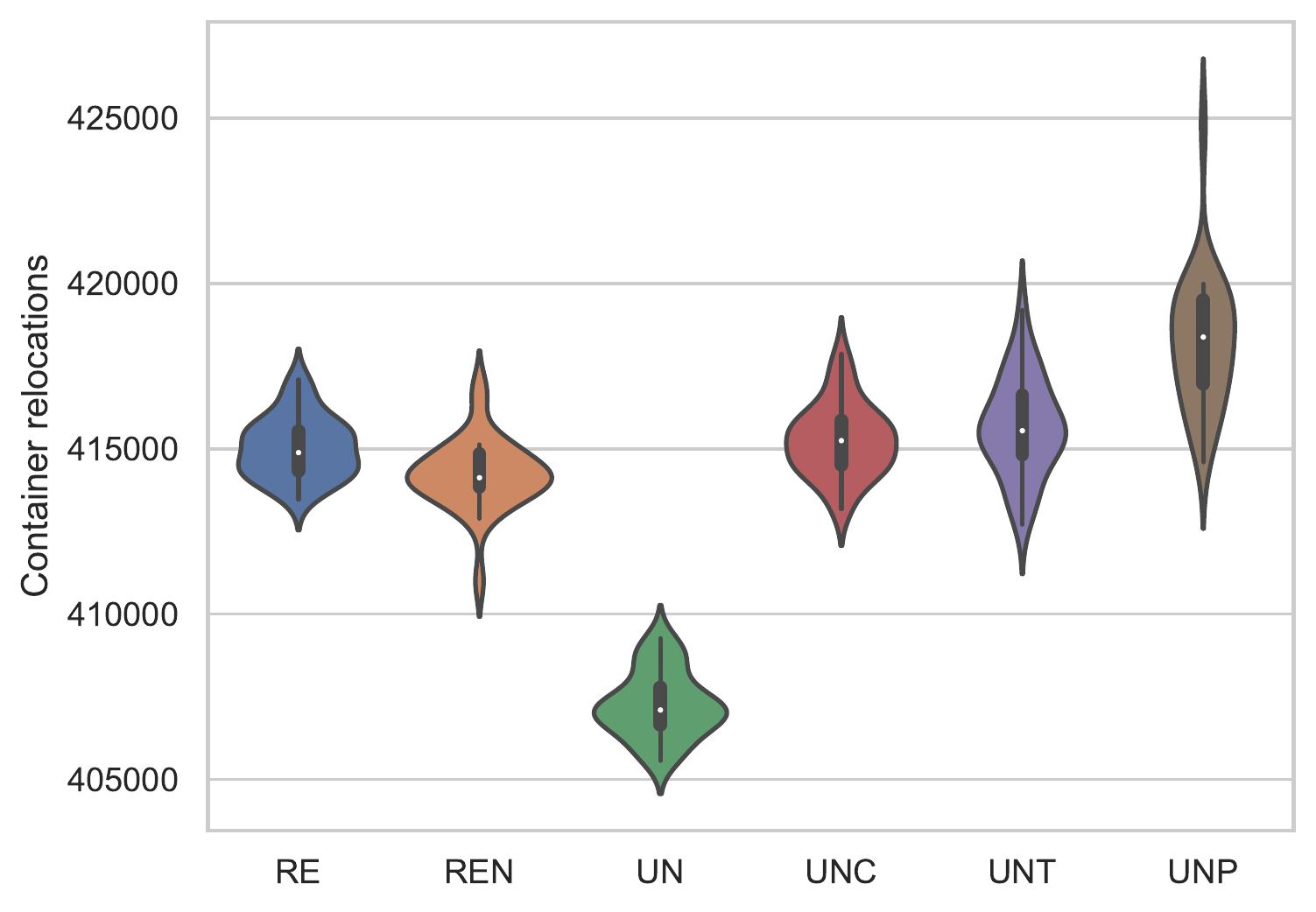}
\caption{Optimising the number of relocations on the Zhu data set}
\label{fig:violinzhurel}
\end{figure}

\begin{figure}[]
\centering
\includegraphics[width=6cm]{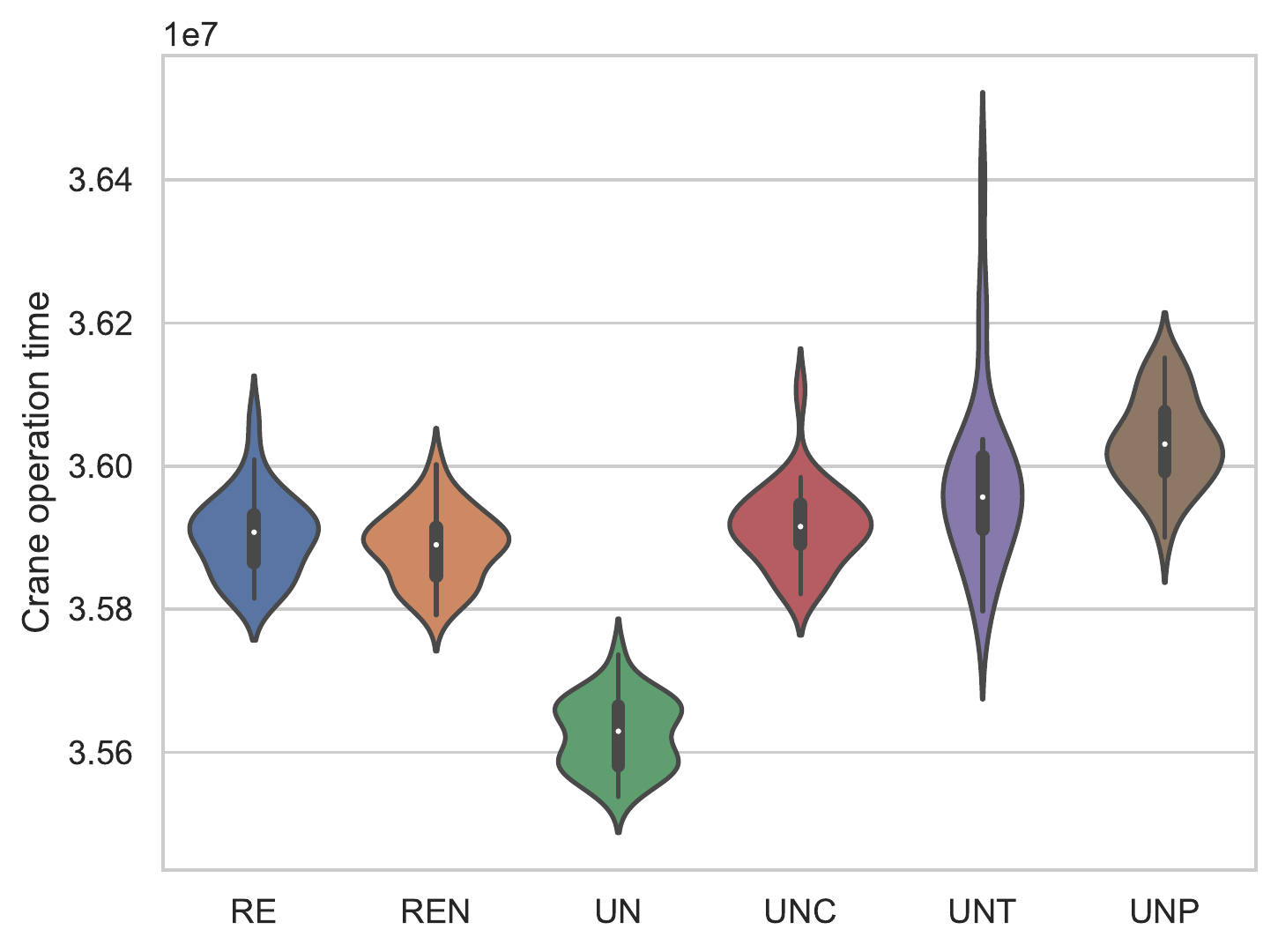}
\caption{Optimising the crane operation time on the Zhu data set}
\label{fig:violinzhutime}
\end{figure}

Figures \ref{fig:violinzhurel} and \ref{fig:violinzhutime} show the violin plots for the tested RSs. Again the UN scheme achieves the best performance. However, in this case, its superiority over the other methods is much more clear. For the other schemes, the conclusions are mostly similar to the Caserta data set. Again, there is no significant difference between the two restricted schemes for both criteria. For the unrestricted schemes, the UN scheme achieves significantly better results than any other scheme.

The previous results demonstrate that the automatically designed RRs almost always achieved a better performance than their manually designed counterparts, almost regardless of the used RS. The results for the different RSs show that certain schemes lead to better results. For the restricted schemes, it was demonstrated that not considering the stack which contains the next target container as in the REN scheme did not significantly improve the results. This suggests that the evolved PFs  can determine on which stack the next target container is located and not relocate containers there if not necessary. Thus, including this strategy in the RS is redundant. For the unrestricted scheme it is clear that evolving two rules which perform different decisions is difficult. It is most appropriate to select the destination container, and use a simple manually designed strategy that determines how to perform the unrestricted moves. 
 
\section{Further analysis}

\subsection{Correlation between the optimised criteria}
\begin{figure}[]
\centering
\includegraphics[width=6cm]{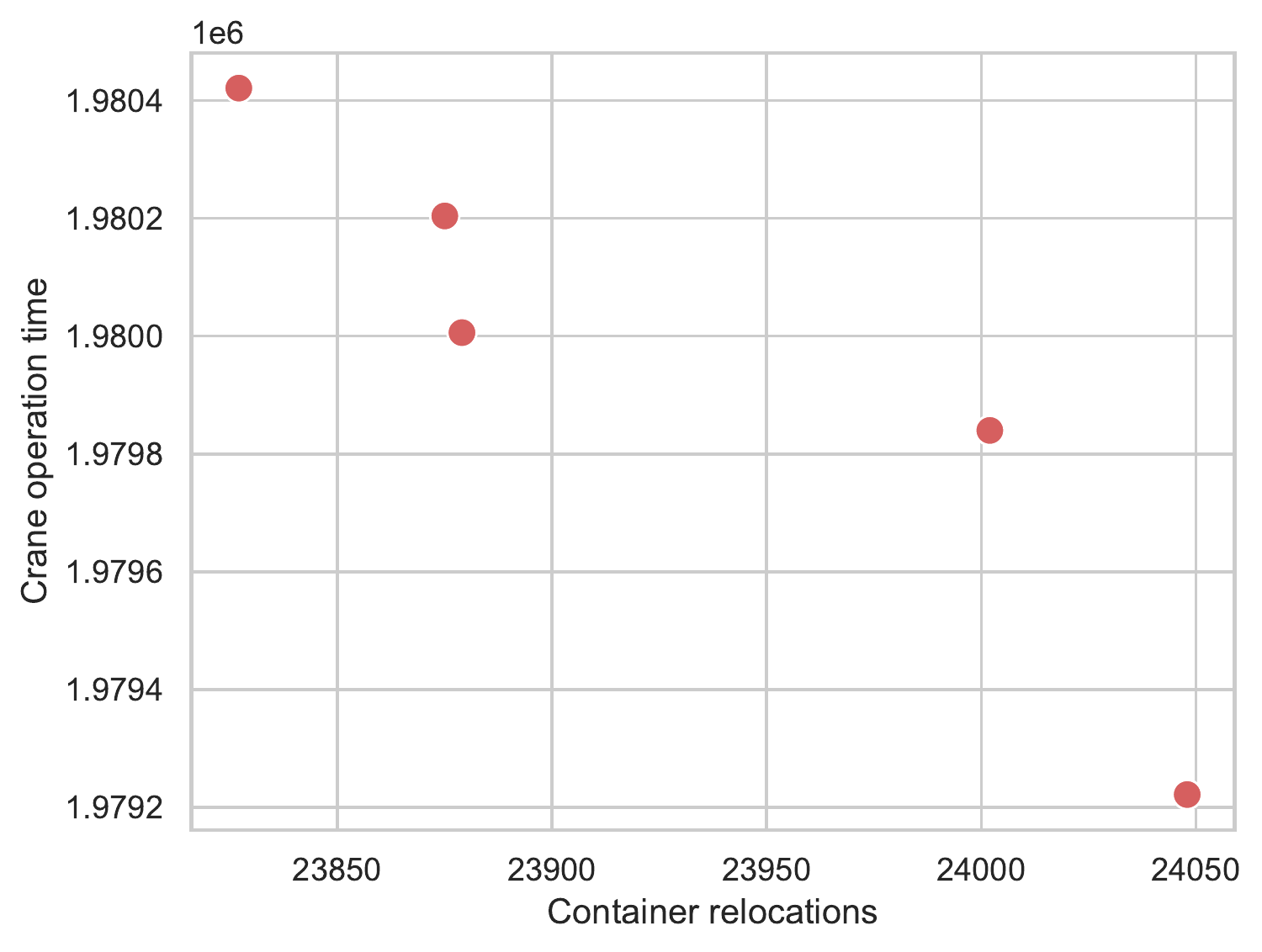}
\caption{Pareto front of solutions when optimising both criteria}
\label{fig:pareto}
\end{figure}

The previous section demonstrated that by optimising one criterion, quite good solutions were obtained for the other criterion, especially when the number of relocations is optimised. This suggests that the two objectives might be positively correlated and that it is enough to optimise only a single objective. To investigate the correlation between the objectives, NSGA-II~\cite{Deb2002} was applied to optimise both criteria simultaneously and obtain Pareto fronts of solutions. The Pareto front obtained for the unrestricted RS is shown in Figure \ref{fig:pareto}. Based on 30 executions, only 5 non-dominated solutions were obtained. The figure shows that the range of the criteria values for these objectives is relatively small, especially when considering the time to perform the relocations. This demonstrates that both criteria are positively correlated and that by optimising the number of container relocations, the crane operation time will be implicitly optimised.  

\subsection{Algorithm convergence}

\begin{figure}[]
\centering
\includegraphics[width=8cm]{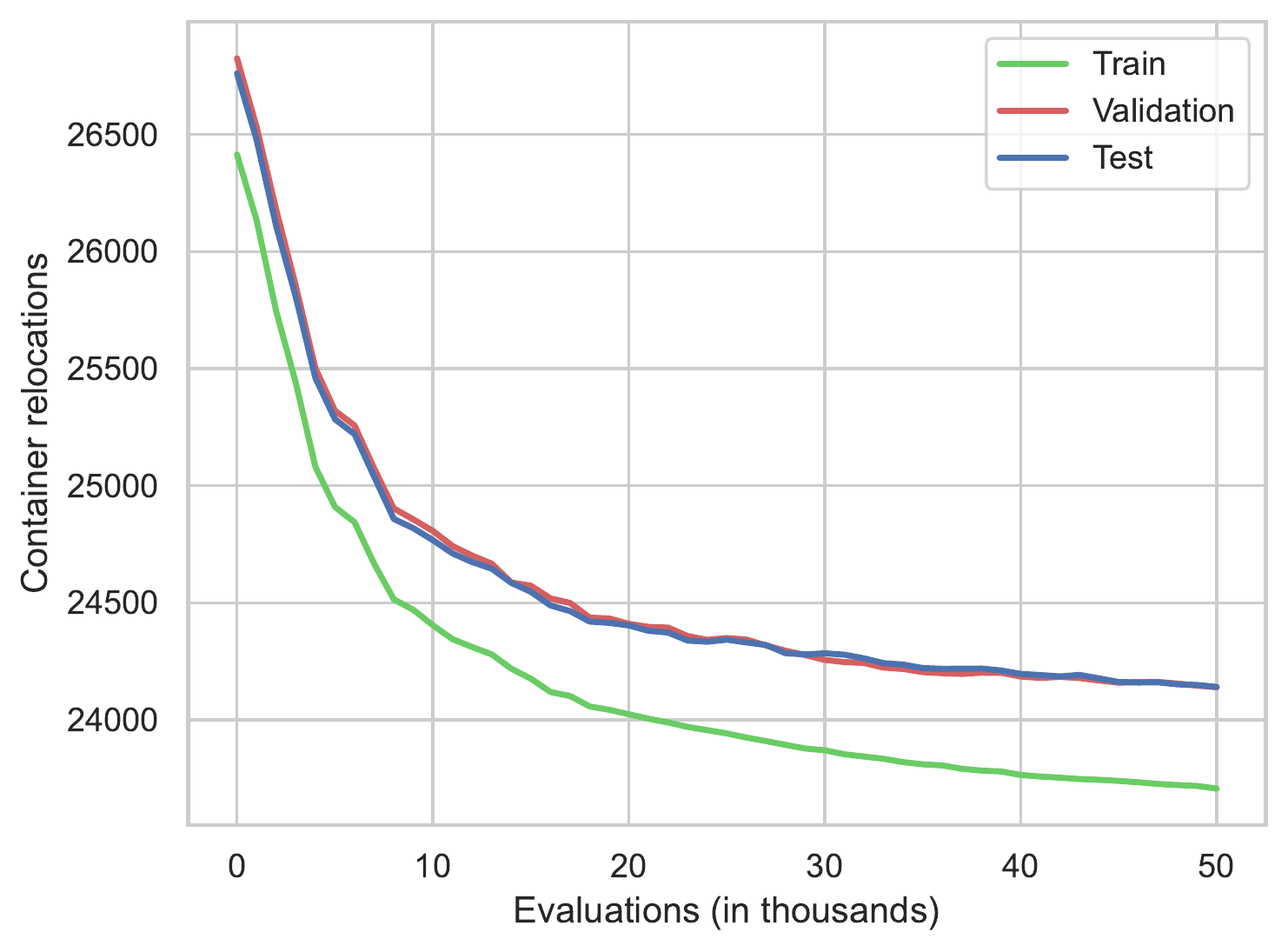}
\caption{Convergence on the problems sets for the restricted variant}
\label{fig:convrest}
\end{figure}

\begin{figure}[]
\centering
\includegraphics[width=8cm]{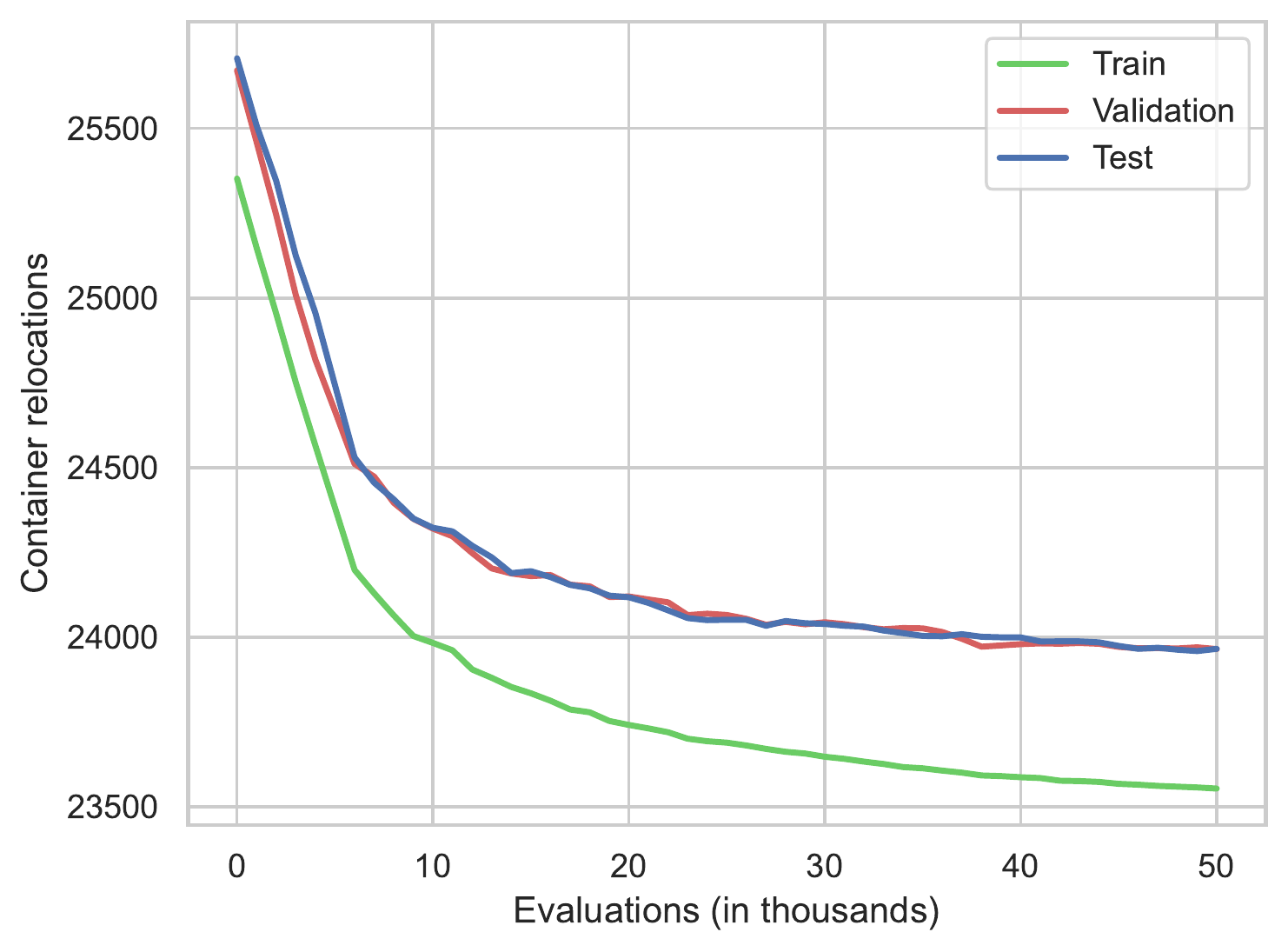}
\caption{Convergence on the problems sets for the unrestricted variant}
\label{fig:convunrest}
\end{figure}

Figures \ref{fig:convrest} and \ref{fig:convunrest} show the convergence of the algorithm on all three datasets for the restricted and unrestricted relocation scheme, respectively. Both figures show that even after 50000 evaluations, the algorithm improves its fitness on the train set. However, on the validation and test set after around 30000 evaluations the fitness slowly starts to stagnate, which denotes that the algorithm has converged to good solutions. An additional thing that can be observed is that there is slight oscillation on the validation and test set during the evolution. This means that the algorithm obtains rules which can generalise well on unseen problem instances. The statistical tests on the validation set show that it would have been possible to stop the algorithm between 30000 and 36000 evaluations without obtaining significantly worse results.

\subsection{Run time analysis}

As with other hyper-heuristic approaches, the runtime of the proposed method can be divided into two parts: the generation and execution of RRs. Since the generation of RRs is performed by GP, this part is more computationally intensive. For the default parameters used, a single run of GP took on average 50 and 40 minutes for the restricted and unrestricted relocation schemes, respectively. However, this part is performed offline at any time to obtain RRs. On the other hand, the exciting time of the evolved RRs is severely smaller. When applied to solve the entire Caserta dataset consisting of 840 instances, the restricted and unrestricted rules require around 0.03 seconds on average. To solve all instances in the Zhu dataset 0.4 seconds were required on average. Therefore, the execution time of the RR is almost negligible, especially for a single decision. 

\subsection{Relocation rule analysis}

\begin{figure}[]
\centering
\includegraphics[width=0.7\columnwidth]{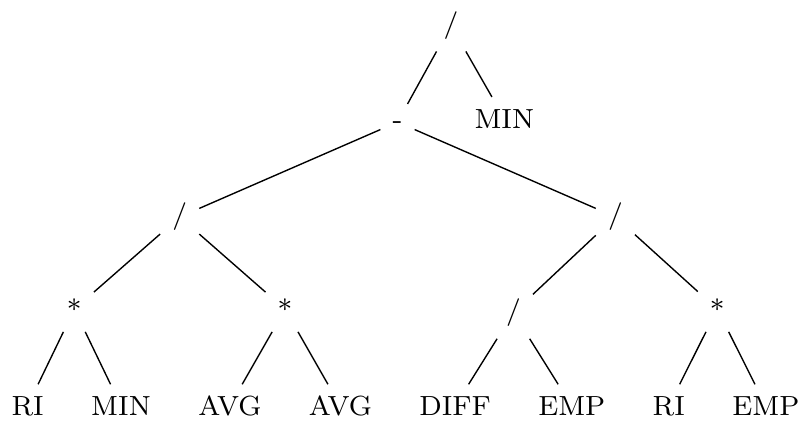}
\caption{Relocation rule example}
\label{fig:ruleexamp}
\end{figure}

To gain a better insight into the workings of the generated RRs, in this section the behaviour of a selected RR for the restricted relocation scheme will be explained. The tree representation of the rule is shown in Figure~\ref{fig:ruleexamp}, which translates to the following expression
$$\frac{\frac{\text{RI}*\text{MIN}}{\text{AVG}*\text{AVG}}-\frac{\text{DIF}}{\text{RI}*\text{EMP}*\text{EMP}}}{\text{MIN}}.$$ 
The first part of the expression $\frac{\text{RI}*\text{MIN}}{\text{AVG}*\text{AVG}}$ favours stacks with a small reshuffle index (containers that have a smaller ID than the container that will be relocated), but which also contain containers with larger IDs. The second expression $-\frac{\text{DIF}}{\text{RI}*\text{EMP}*\text{EMP}}$ favours stacks with a large difference between the container with the minimum ID on the stack and the one that is being relocated, but which also contain as few free spaces as possible. Thus, the second part prefers stacks that are almost full but have a small reshuffle index, and thus the number of unnecessary relocations is minimised. In the end, the rules try to balance between the stacks that have a small reshuffle index for the current container and are as full as possible, and have a high average of IDs, and with that, it tries to minimise the number of relocations. 

\subsection{Analysis of the priority function sizes}
The size of the evolved priority functions is an essential factor when designing RRs. The evolved PFs should be small in size so that they are easy to interpret and understand. Unfortunately, GP is prone to bloating, and the evolved expressions will usually contain many redundant parts. Although different methods can be used to remedy this problem, the easiest way is to restrict the depth of the evolved trees. Therefore, the influence of different tree depths on the quality and size of the evolved PFs are explored. Tree depths between 3 and 10 were used, and the obtained results are outlined in Table~\ref{tbl:treedeptch}. The SOB column represents the number of nodes in the expression of the best solution, whereas the MES column denotes the median size of the evolved expressions. The smallest tree depth of 3 and the largest depths from 8 until 10 usually lead to worse results. These depths offer either a too small or too large search space. The best results are usually achieved for depths between 4 and 6. Regarding the tree sizes, it is clear that a certain tree depth is required to obtain good solutions. For example, the median tree size for the depth of 3 is equal to the maximum number of nodes in the tree. This means that this size does not offer enough space for GP to obtain good solutions since better solutions usually contain at least around 30 nodes. Therefore, a tree depth of at least 4 should be used, or even larger depths if it is required to obtain better results. 

\begin{table}[]
\centering
\caption{Influence of the tree depth on rule sizes}
\label{tbl:treedeptch}
\begin{tabular}{@{}lcccc|cc@{}}
\toprule
                    &       & \multicolumn{3}{c|}{Container relocations} & \multicolumn{2}{c}{Size} \\ 
                    & Depth & Min       & Med      & Max      & SOB         & MES        \\ \midrule
\multirow{8}{*}{RE} & 3     & 24230     & 24512    & 24743    & 13          & 15         \\
                    & 4     & 23898     & 24297    & 24615    & 27          & 25         \\
                    & 5     & 23934     & 24219    & 24605    & 41          & 49         \\
                    & 6     & 23808     & 24113    & 24531    & 81          & 66         \\
                    & 7     & 23907     & 24148    & 24610    & 61          & 90         \\
                    & 8     & 23906     & 24238    & 25576    & 191         & 177        \\
                    & 9     & 23816     & 24285    & 25060    & 409         & 180        \\
                    & 10    & 23956     & 24286    & 25219    & 93          & 241        \\ \midrule
\multirow{8}{*}{UN} & 3     & 24002     & 24230    & 24444    & 15          & 15         \\
                    & 4     & 23862     & 24032    & 24406    & 31          & 27         \\
                    & 5     & 23679     & 23967    & 24122    & 61          & 46         \\
                    & 6     & 23818     & 23989    & 24407    & 103         & 62         \\
                    & 7     & 23731     & 24009    & 24303    & 111         & 99         \\
                    & 8     & 23809     & 24082    & 24447    & 65          & 87         \\
                    & 9     & 23852     & 24144    & 24648    & 37          & 156        \\
                    & 10    & 23835     & 24151    & 24630    & 39          & 247        \\ \bottomrule
\end{tabular}
\end{table}

\begin{figure}[]
\includegraphics[width=8cm]{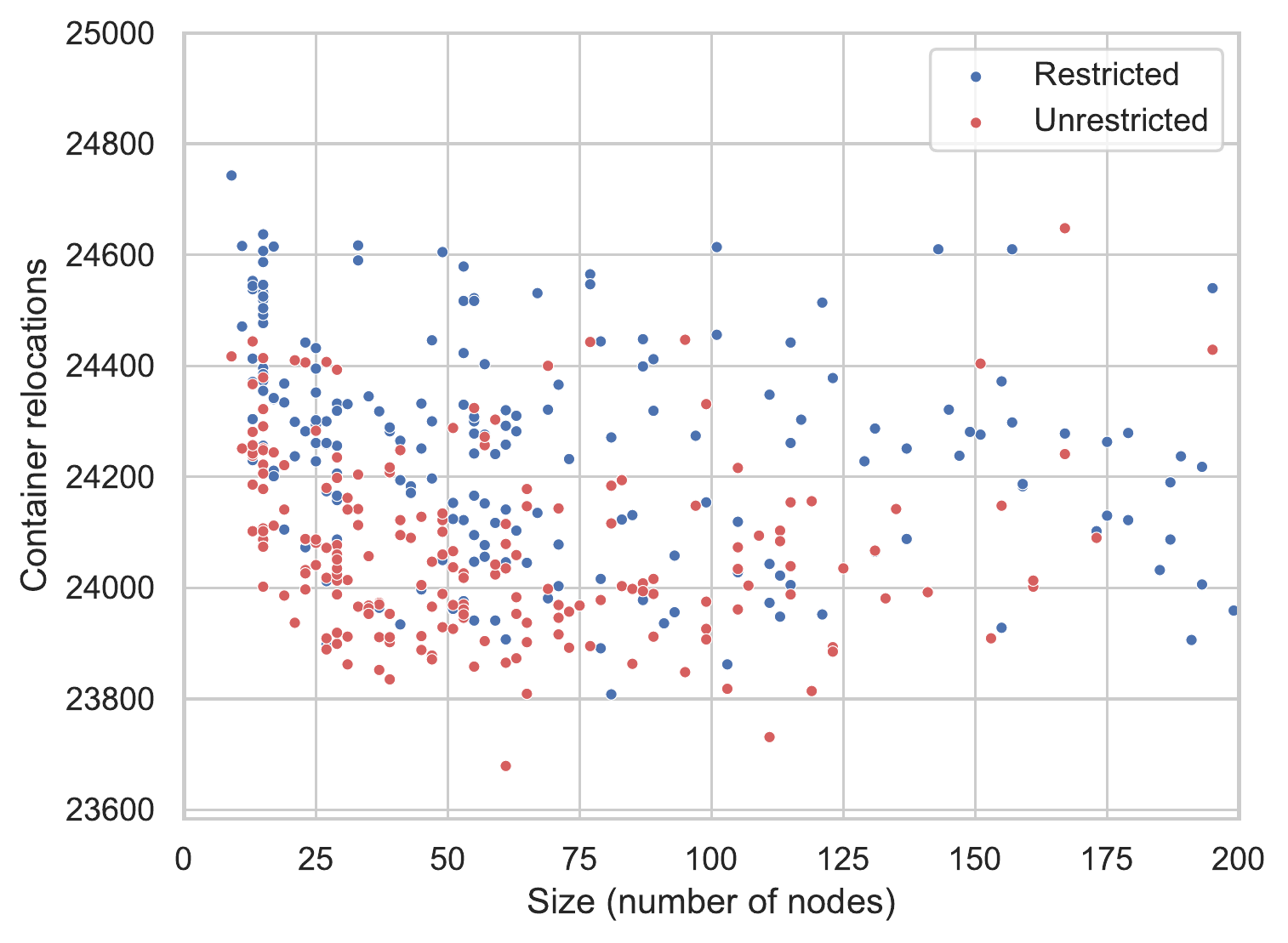}
\caption{Correlation between the fitness of the rules and expression size}
\label{fig:relsize}
\end{figure}

To gain a better understanding on the correlation of the tree sizes and their fitness, all solutions obtained for the tested tree depths were plotted against their size and fitness in Figure~\ref{fig:relsize}. It should be noted that the tree sizes were cut off at 200 nodes to make the figure more readable. This figure shows that until a specific tree size, it is not possible to obtain good expressions. Although it is difficult to determine a fixed size at which this happens, the figure shows that after tree sizes of 25 nodes, the number of good solutions increases. The increase of tree size does not guarantee that better solutions are obtained, but it does increase the probability of it. 

\subsection{Generality of rules}

In the previous section, the RRs were applied on the data set of the same type as the one on which they were evolved. Naturally, the best results should be obtained if a data set used to generate the rules is similar to the one on which the rules will be applied. However, the obtained rules should be general enough and perform well on other types of data sets as well. To test this, the rules that were evolved on the Caserta dataset were tested on the Zhu data set and vice versa. Table~\ref{tbl:generality} outlines the results obtained when such a switch is performed. The first 4 rows show the results obtained when the Caserta data set is used for testing, whereas the last 4 rows denote the results obtained when the Zhu dataset is used as the test set. The postfixes "-C" and "-Z" denote that the rules were trained on the Caserta and Zhu data set types, respectively. 

When using the Caserta data set type as the test set, an interesting thing can be observed. For the unrestricted variant, there is no significant difference between the rules generated using either dataset. However, for the restricted variant, the rules generated using the Zhu dataset perform significantly worse than those generated on the Caserta data set. On the other hand, when the Zhu data set is used to test the rules generated using the Zhu and Caserta data sets, it performs similarly for both relocation schemes. This seems to suggest that using the Caserta data set for training might result in more general rules. A possible reason could be that the yards in the Zhu instances are almost full, limiting the relocation possibilities. This might make finding good rules more difficult as there are only a few possible relocations in the first steps, and thus the rules have to focus more on the latter part when there are more available places. 

\begin{table}[]
\centering
\caption{Results when the RRs are applied on the other data set type}
\label{tbl:generality}
\adjustbox{max width=\columnwidth}{%
\begin{tabular}{@{}llccc|ccc@{}}
\toprule
                         &      & \multicolumn{3}{c}{Container relocations}                                             & \multicolumn{3}{c}{Crane operation time}                                                    \\ 
                         &      & \multicolumn{1}{c}{Min} & \multicolumn{1}{c}{Med} & \multicolumn{1}{c}{Max} & \multicolumn{1}{c}{Min} & \multicolumn{1}{c}{Med} & \multicolumn{1}{c}{Max} \\ \midrule
\multirow{4}{*}{\rotatebox[origin=c]{90}{Caserta}} & RE-C & 23816                   & 24122                   & 24508                   & 1978200                 & 1993387                 & 2004644                 \\
                         & RE-Z & 24281                   & 24871                   & 25600                   & 2000882                 & 2019038                 & 2052329                 \\
                         & UN-C & 23679                   & 23967                   & 24122                   & 1975008                 & 1985987                 & 1996640                 \\
                         & UN-Z & 23754                   & 24057                   & 24673                   & 1979928                 & 1991150                 & 2018590                 \\ \midrule
\multirow{4}{*}{\rotatebox[origin=c]{90}{Zhu}}    & Re-Z & 413469                  & 414885                  & 417092                  & 35920974                & 35989754                & 36228750                \\
                         & RE-C & 412559                  & 416128                  & 422553                  & 35868280                & 36011830                & 36449254                \\
                         & UN-Z & 405571                  & 407103                  & 409266                  & 35625504                & 35681901                & 35926484                \\
                         & UN-C & 406239                  & 407828                  & 411046                  & 35658306                & 35706072                & 35967755                \\ \bottomrule
\end{tabular}}
\end{table}

\subsection{Performance on large problems}

To stress test the proposed approach, the RRs evolved in the previous section when optimising the number of relocations for the Caserta dataset are applied on a large scale Caserta data set that was proposed in~\cite{Tricoire2018}. This data set is composed of 400 instances that contain from 10 to 100 stacks and 100 containers per stack, meaning that the largest instances contain 10000 containers. The results obtained on this data set are denoted in Table~\ref{tbl:large}. The table shows that the automatically designed RRs significantly outperform all the manually designed rules. The median value of the automatically designed rules is better by around 30\% when compared with the best manually designed rule PU2. This shows that the gap between the automatically and manually designed rules increases as the instances become larger. Since the automatically designed rules were evolved on the small scale Caserta data set, this result also demonstrates that the obtained rules generalise well on unseen instances.

\begin{table}[]
\centering
\caption{Results for large scale Caserta dataset}
\label{tbl:large}
\adjustbox{max width=\columnwidth}{%
\begin{tabular}{@{}lcccccc@{}}
\toprule
    & \multicolumn{3}{c}{Container relocations}                                                 & \multicolumn{3}{c}{Crane operation time}     \\ \midrule
TLP & \multicolumn{3}{c}{29717794}                                                    & \multicolumn{3}{c}{3.1E+9} \\
RI  & \multicolumn{3}{c}{15104213}                                                    & \multicolumn{3}{c}{1.6E+9} \\
MM  & \multicolumn{3}{c}{18902574}                                                    & \multicolumn{3}{c}{2.0E+9} \\
PR3 & \multicolumn{3}{c}{11119625}                                                    & \multicolumn{3}{c}{1.3E+9} \\
PR4 & \multicolumn{3}{c}{10966282}                                                    & \multicolumn{3}{c}{1.3E+9} \\
PU1 & \multicolumn{3}{c}{10951307}                                                    & \multicolumn{3}{c}{1.3E+9} \\
PU2 & \multicolumn{3}{c}{10797383}                                                    & \multicolumn{3}{c}{1.2E+9} \\ \midrule
    & Min                         & \multicolumn{1}{c}{Med} & \multicolumn{1}{c}{Max} & Min      & Med     & Max     \\ \midrule
RE  & \multicolumn{1}{l}{5212033} & 7304279                 & 10657770                & 6.9E+8   & 8.9E+8  & 1.2E+9  \\
UN  & \multicolumn{1}{l}{5720948} & 7370982                 & 9444975                 & 7.4E+8   & 9.2E+8  & 1.1E+9  \\ \bottomrule
\end{tabular}}
\end{table}

\subsection{Improving results by evolving two PFs}
During preliminary analyses, it was observed that frequently a single rule had difficulties performing well across a broader range of instances. This naturally raises the question of whether evolving several expressions and using them simultaneously can improve the results. Therefore, the RE and UN schemes were extended to decide where to relocate the current container based on two PFs. Both expressions are evaluated individually for each candidate stack, and the current container will be relocated to the stack, which obtained the minimum priority by any of the two expressions. 

\begin{figure}[!tbh]  
     \centering
     \subfloat[Caserta dataset]{
         \centering
         \includegraphics[width=0.24\textwidth]{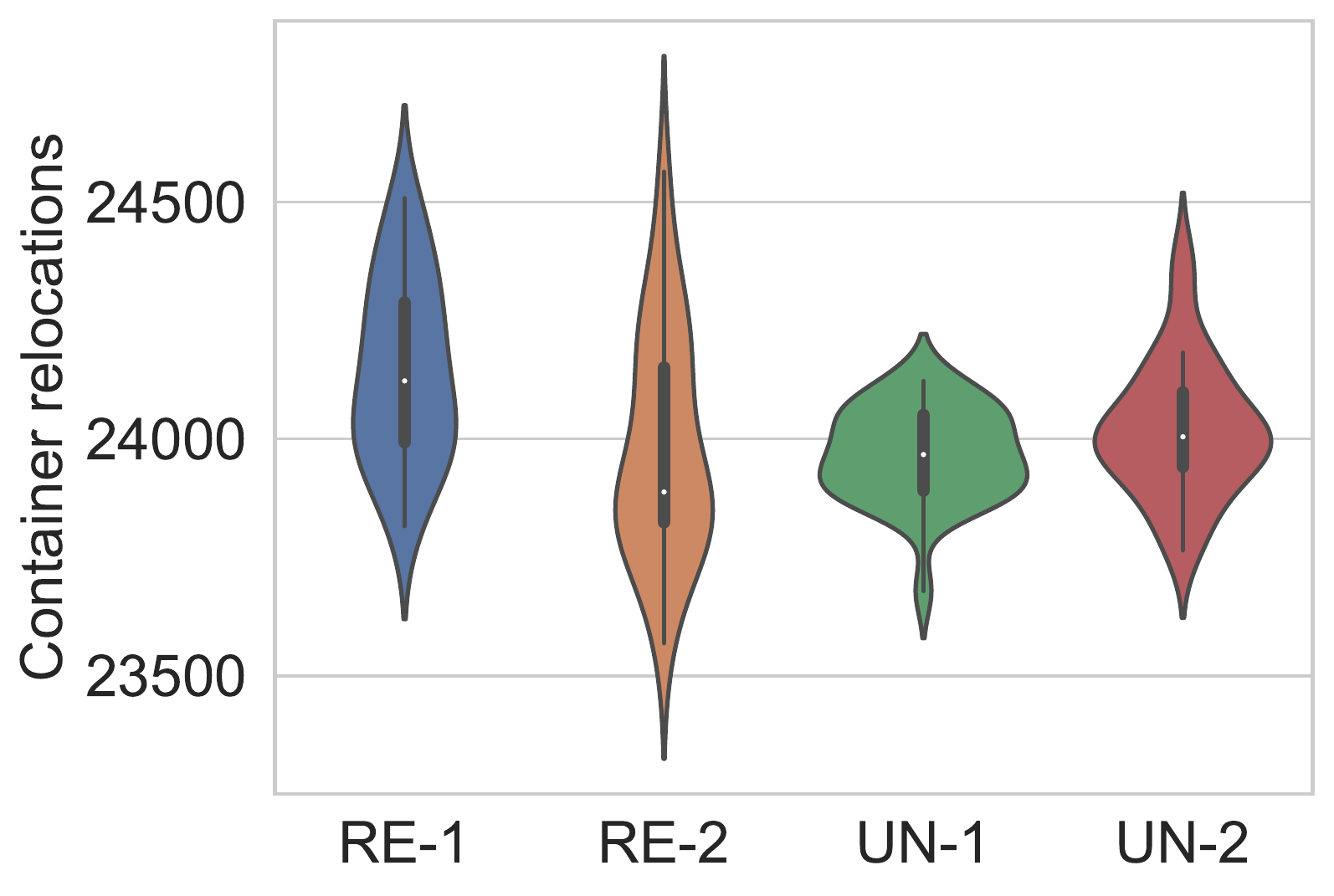}
         \label{fig:roulette}
     }
    \subfloat[Zhu dataset]{
         \centering
         \includegraphics[width=0.24\textwidth]{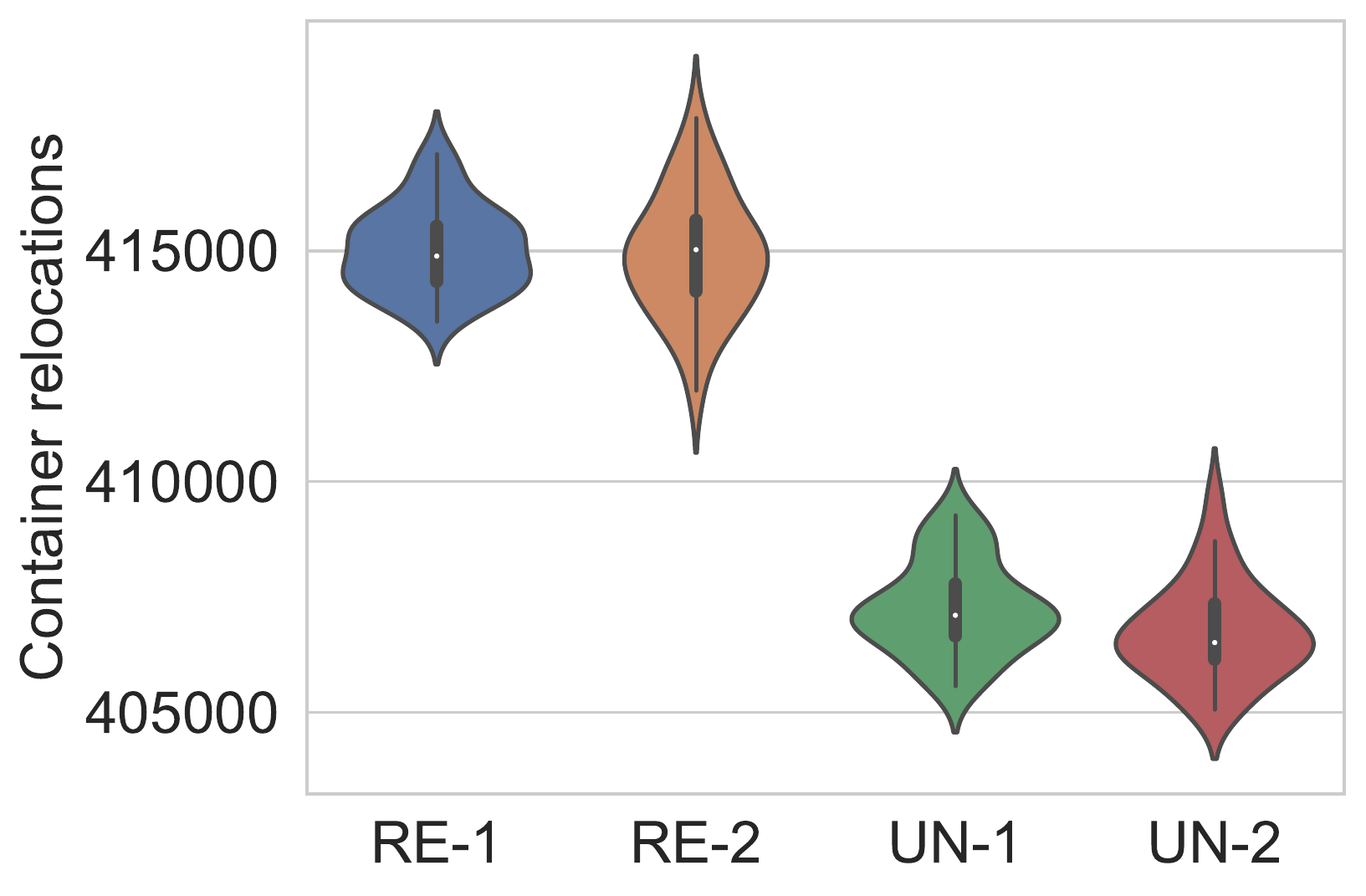}
         \label{fig:roulette_lexi}
     }
    \caption{Violin plots of RRs using one and two PFs}
    \label{fig:tworules}
\end{figure}

The results for this approach (denoted with the suffix "-2") are compared to the standard single expression RRs (denoted with the suffix "-1") are denoted in Table~\ref{tbl:tworules} and Figure~\ref{fig:tworules}. The results demonstrate that using two PFs has an effect on the obtained results, however, more in the case when using the RE scheme. This is especially evident for the Caserta data set, on which RRs that use 2 PFs significantly outperform the version that uses only a single PF. Not only that, in this case, the RRs using the RE scheme obtained a better median than the rules using the UN scheme. Furthermore, RE-2 achieved the best overall results across all the performed experiments. This clearly shows that the RE scheme is not inferior to the UN scheme. The reason it performed worse than the UN scheme could have been because a single PF was not enough to incorporate all the relevant information to perform such good decisions. Using two rules gives GP the ability that each one specialises for different situations. These results provide motivation for further research in which ensemble learning methods could be applied to generate sets of RRs.

\begin{table}[]
\centering
\caption{RRs with two expressions}
\label{tbl:tworules} 
\adjustbox{max width=\columnwidth}{%
\begin{tabular}{@{}llccc|ccc@{}}
\toprule
                         &      & \multicolumn{3}{c|}{Container relocations} & \multicolumn{3}{c}{Crane operation time}       \\ 
                         &      & Min       & Med      & Max      & Min      & Med      & Max      \\ \midrule
\multirow{4}{*}{\rotatebox[origin=c]{90}{Caserta}} & RE-1 & 23816     & 24122    & 24508    & 1978200  & 1993387  & 2004644  \\
                         & RE-2 & 23569     & 23888    & 24564    & 1972846  & 1983899  & 2015048  \\
                         & UN-1 & 23679     & 23967    & 24122    & 1975008  & 1985987  & 1996640  \\
                         & UN-2 & 23765     & 24005    & 24377    & 1978301  & 1991808  & 2001211  \\ \midrule
\multirow{4}{*}{\rotatebox[origin=c]{90}{Zhu}}     & RE-1 & 413469    & 414885   & 417092   & 35920974 & 35989754 & 36228750 \\
                         & RE-2 & 411973    & 415025   & 417873   & 35888624 & 35986230 & 36231655 \\
                         & UN-1 & 405571    & 407103   & 409266   & 35625504 & 35681901 & 35926484 \\
                         & UN-2 & 405062    & 406512   & 409652   & 35623466 & 35688671 & 35894462 \\ \bottomrule
\end{tabular}}
\end{table}

\subsection{Terminal set analysis}
\begin{figure}[]
\centering
\includegraphics[width=7cm]{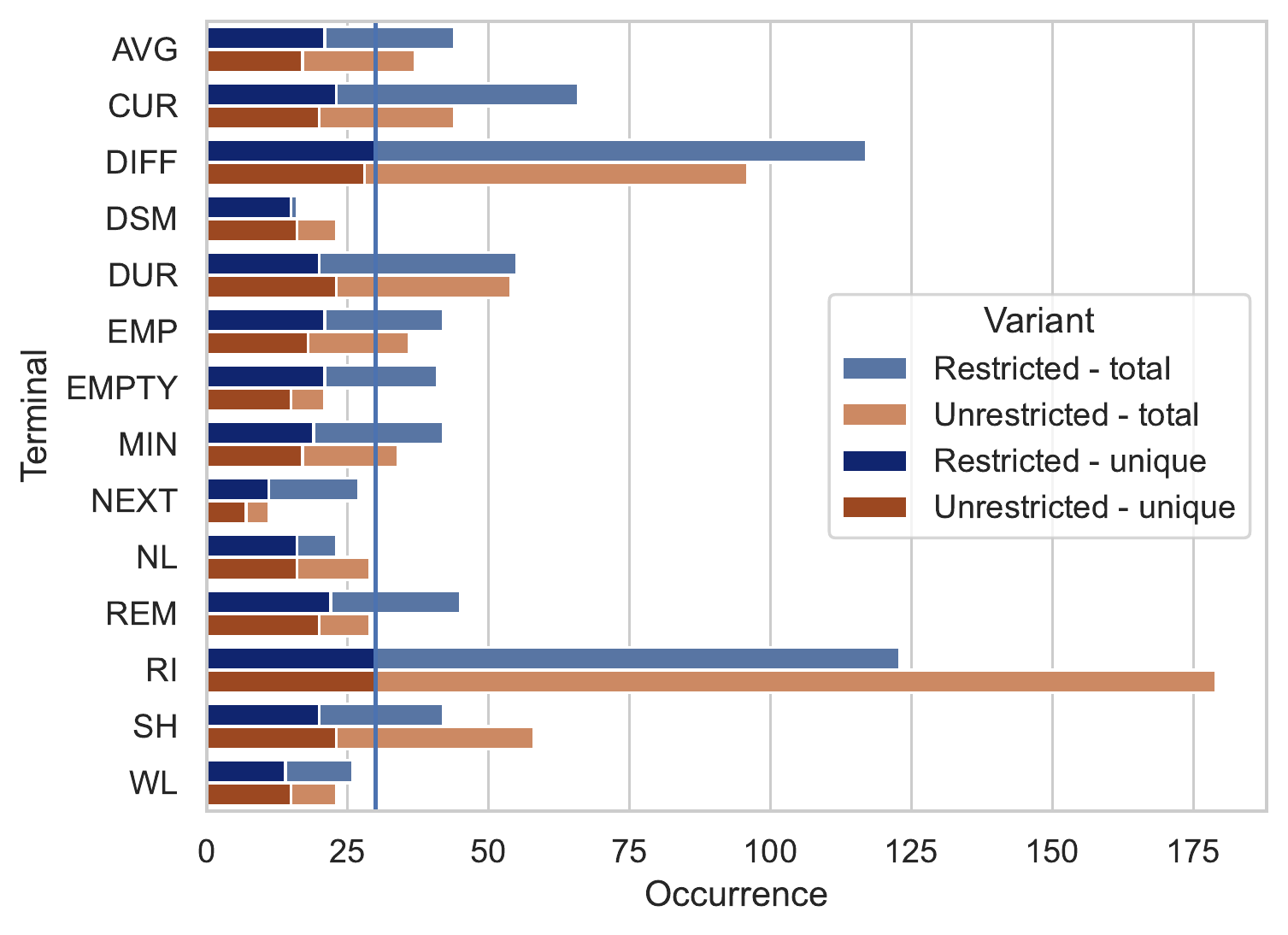}
\caption{Terminal node occurrence frequency in the evolved PFs}
\label{fig:nodeocc}
\end{figure}

In the proposed method, 14 terminal nodes, which encompass the characteristics of the problem, were proposed. Although using all those in the construction of PFs leads to good results, an open question is whether all those nodes are required. To gain a better insight into which nodes are the most important, a short analysis is performed. First, based on the initial experiments for the RE and UN, the occurrences of all nodes were calculated and are denoted in Figure~\ref{fig:nodeocc}. The figure shows the total number of times that a particular node occurred in the trees (marked as total) and the unique number of times it occurred in an expression (denoted as unique), meaning that the number of expressions that contain a particular node is counted. For the latter metric, the maximum value is equal to the number of runs, meaning 30. Therefore, a vertical blue line was plotted at that value to denote better which nodes appear in most individuals. The figure denotes that two terminals are clearly more important than others. The two most important terminals are the RI and DIFF terminals. They appear not only in almost all generated rules but also appear several times in them. This is expected because they give direct information about whether it makes sense to relocate the current container to a stack. Out of the other nodes, the CUR, DUR, and SH also appear quite often, but not with such a large frequency. The least informative nodes are DSM, NEXT, NL, and WL.

Based on the initial observation that certain terminal nodes are less important, a series of experiments were performed to decrease the number of terminals in the set. At the start, the set contained all terminals, and then the node, which, when removed from the set, leads to the best results on the validation set, is excluded from the terminal set. The procedure is then repeated until all the nodes are removed. Although not all node combinations are tested (which would be hardly possible due to the sheer number of combinations), such a procedure still leads to a reduced set of terminal nodes. For the RE scheme, the terminal set was reduced to six nodes: SH, RI, EMP, DIFF, AVG, and CUR. On the other hand, in the UN scheme, the best results are obtained using only 4 nodes, SH, EMP, DIFF, and RI. The results obtained by the reduced terminal sets are denoted in Table~\ref{tbl:reduced}, and are denoted with the suffix "-R". For both schemes, it was possible to improve the performance in comparison with the original terminal set. Although the differences to the original set are not large, the results obtained with the reduced terminal set were significantly better. 

\begin{table}[]
\centering
\caption{Results of RRs with a reduced terminal set}
\label{tbl:reduced}
\adjustbox{max width=\columnwidth}{%
\begin{tabular}{@{}llccc|ccc@{}}
\toprule
                         &      & \multicolumn{3}{c|}{Container relocations} & \multicolumn{3}{c}{Crane operation time}       \\ 
                         &      & Min       & Med      & Max      & Min      & Med      & Max      \\ \midrule
\multirow{4}{*}{\rotatebox[origin=c]{90}{Caserta}} & RE & 23816     & 24122    & 24508    & 1978200  & 1993387  & 2004644  \\
                         & RE-R & 23758     & 23983    & 24252    & 1974572  & 1983939  & 1994616  \\
                         & UN & 23679     & 23967    & 24122    & 1975008  & 1985987  & 1996640  \\
                         & UN-R & 23786     & 23901    & 24126    & 1978344  & 1981985  & 1990452  \\ \midrule
\end{tabular}}
\end{table}

\section{Conclusion}

This study proposed the application of GP to design RRs for the CRP automatically. In the proposed method, GP is used to evolve an expression that assigns priorities to stacks based on which it is determined on which stack the container will be placed. An RS is then used to relocate and retrieve containers from the yard using the expression generated by GP. The goal of this approach was to generate RRs which perform better than existing manually designed rules from the literature.

Experimental results demonstrate that automatically designed RRs outperform several existing rules from the literature for the two considered criteria. Even more, it was demonstrated that even the worst automatically designed rules perform better than their manually designed counterparts for the best relocation schemes. The additional analysis further showed that the generated RRs are general enough to perform well on problems of different properties than those they were evolved on, that the training time can be further reduced without a significant impact on the obtained results, and that their performance can be improved by using a subset of terminal nodes or evolving two PFs for container ranking. The obtained results and observations demonstrate that the proposed method represents a viable alternative to manually designed RRs, which can be utilised by themselves to construct good solutions in a relatively small amount of time. Additionally, because of their low execution time, they also have a potential of being used in synergy with other methods similarly as manually designed RRs have been used. 

In future work, it is planned to extend this work in several directions. Firstly, the method will be adapted and tested on other variants of the CRP, like the multibay and duplicate priority variants. Another direction is to improve the obtained results further using ensemble learning methods or the rollout heuristics. Finally, the last research direction would be to adapt the proposed method and apply it to the dynamic variant of the CRP problem, in which the sequence of container retrievals is not entirely known beforehand, and the order of retrievals becomes known during the execution of the system.

\ifCLASSOPTIONcaptionsoff
  \newpage
\fi



\bibliographystyle{IEEEtran}
%
\bibliography{bare_jrnl}

\begin{thebibliography}{10}
\providecommand{\url}[1]{#1}
\csname url@samestyle\endcsname
\providecommand{\newblock}{\relax}
\providecommand{\bibinfo}[2]{#2}
\providecommand{\BIBentrySTDinterwordspacing}{\spaceskip=0pt\relax}
\providecommand{\BIBentryALTinterwordstretchfactor}{4}
\providecommand{\BIBentryALTinterwordspacing}{\spaceskip=\fontdimen2\font plus
\BIBentryALTinterwordstretchfactor\fontdimen3\font minus
  \fontdimen4\font\relax}
\providecommand{\BIBforeignlanguage}[2]{{%
\expandafter\ifx\csname l@#1\endcsname\relax
\typeout{** WARNING: IEEEtran.bst: No hyphenation pattern has been}%
\typeout{** loaded for the language `#1'. Using the pattern for}%
\typeout{** the default language instead.}%
\else
\language=\csname l@#1\endcsname
\fi
#2}}
\providecommand{\BIBdecl}{\relax}
\BIBdecl

\bibitem{Jovanovic2019}
R.~Jovanovic, S.~Tanaka, T.~Nishi, and S.~Vo{\ss}, ``{A GRASP approach for
  solving the Blocks Relocation Problem with Stowage Plan},'' \emph{Flexible
  Services and Manufacturing Journal}, vol.~31, no.~3, pp. 702--729, 2019.

\bibitem{Steenken2004}
D.~Steenken, S.~Vo{\ss}, and R.~Stahlbock, ``{Container terminal operation and
  operations research - A classification and literature review},'' \emph{OR
  Spectrum}, vol.~26, no.~1, pp. 3--49, 2004.

\bibitem{Lu2020}
C.~Lu, B.~Zeng, and S.~Liu, ``{A Study on the Block Relocation Problem: Lower
  Bound Derivations and Strong Formulations},'' \emph{IEEE Transactions on
  Automation Science and Engineering}, pp. 1--25, 2020.

\bibitem{SCULLI1988585}
D.~Sculli and C.~Hui, ``Three dimensional stacking of containers,''
  \emph{Omega}, vol.~16, no.~6, pp. 585--594, 1988.

\bibitem{Avriel2000}
M.~Avriel, M.~Penn, and N.~Shpirer, ``Container ship stowage problem:
  complexity and connection to the coloring of circle graphs,'' \emph{Discrete
  Applied Mathematics}, vol. 103, no. 1-3, pp. 271--279, Jul. 2000.

\bibitem{Kim2006}
K.~H. Kim and G.~P. Hong, ``{A heuristic rule for relocating blocks},''
  \emph{Computers and Operations Research}, vol.~33, no.~4, pp. 940--954, 2006.

\bibitem{Wu2010}
K.-C. Wu and C.-J. Ting, ``{A beam search algorithm for minimizing reshuffle
  operations at container yards},'' pp. 703--710, 2010.

\bibitem{Diaz2017}
C.~D{\'{i}}az and M.~C. Riff, ``{New bounds for large container relocation
  instances using grasp},'' \emph{Proceedings - 2016 IEEE 28th International
  Conference on Tools with Artificial Intelligence, ICTAI 2016}, pp. 343--349,
  2017.

\bibitem{Lee2010}
Y.~Lee and Y.~J. Lee, ``{A heuristic for retrieving containers from a yard},''
  \emph{Computers and Operations Research}, vol.~37, no.~6, pp. 1139--1147,
  2010.

\bibitem{Caserta2011}
M.~Caserta, S.~Vo{\ss}, and M.~Sniedovich, ``{Applying the corridor method to a
  blocks relocation problem},'' \emph{OR Spectrum}, vol.~33, no.~4, pp.
  915--929, 2011.

\bibitem{Caserta2012}
M.~Caserta, S.~Schwarze, and S.~Vo{\ss}, ``{A mathematical formulation and
  complexity considerations for the blocks relocation problem},''
  \emph{European Journal of Operational Research}, vol. 219, no.~1, pp.
  96--104, 2012.

\bibitem{Forster2012}
F.~Forster and A.~Bortfeldt, ``{A tree search procedure for the container
  relocation problem},'' \emph{Computers and Operations Research}, vol.~39,
  no.~2, pp. 299--309, 2012.

\bibitem{Zhu2012}
W.~Zhu, H.~Qin, A.~Lim, and H.~Zhang, ``{Iterative deepening A* algorithms for
  the container relocation problem},'' \emph{IEEE Transactions on Automation
  Science and Engineering}, vol.~9, no.~4, pp. 710--722, 2012.

\bibitem{Jovanovic2014}
R.~Jovanovic and S.~Vo{\ss}, ``{A chain heuristic for the blocks relocation
  problem},'' \emph{Computers and Industrial Engineering}, vol.~75, no.~1, pp.
  79--86, 2014.

\bibitem{LIN2015132}
D.-Y. Lin, Y.-J. Lee, and Y.~Lee, ``The container retrieval problem with
  respect to relocation,'' \emph{Transportation Research Part C: Emerging
  Technologies}, vol.~52, pp. 132--143, 2015.

\bibitem{Borjian2015}
S.~Borjian, V.~Galle, V.~H. Manshadi, C.~Barnhart, and P.~Jaillet, ``{Container
  Relocation Problem: Approximation, Asymptotic, and Incomplete Information},''
  2015.

\bibitem{KU20161031}
\BIBentryALTinterwordspacing
D.~Ku and T.~S. Arthanari, ``Container relocation problem with time windows for
  container departure,'' \emph{European Journal of Operational Research}, vol.
  252, no.~3, pp. 1031--1039, 2016. [Online]. Available:
  \url{https://www.sciencedirect.com/science/article/pii/S0377221716001016}
\BIBentrySTDinterwordspacing

\bibitem{Maglic2019}
L.~Magli{\'{c}}, M.~Guli{\'{c}}, and L.~Magli{\'{c}}, ``{Optimization of
  Container Relocation Operations in Port Container Terminals},''
  \emph{Transport}, vol.~35, no.~1, pp. 37--47, 2019.

\bibitem{Cifuentes2020}
C.~D. Cifuentes and M.~C. Riff, ``{G-CREM: A GRASP approach to solve the
  container relocation problem for multibays},'' \emph{Applied Soft Computing
  Journal}, no. xxxx, p. 106721, 2020.

\bibitem{Raggl2020}
S.~Raggl, A.~Beham, S.~Wagner, and M.~Affenzeller, ``{Solution approaches for
  the dynamic stacking problem},'' \emph{GECCO 2020 Companion - Proceedings of
  the 2020 Genetic and Evolutionary Computation Conference Companion}, pp.
  1652--1660, 2020.

\bibitem{Burke2013}
E.~K. Burke, M.~Gendreau, M.~Hyde, G.~Kendall, G.~Ochoa, E.~\"{O}zcan, and
  R.~Qu, ``Hyper-heuristics: a survey of the state of the art,'' \emph{Journal
  of the Operational Research Society}, vol.~64, no.~12, pp. 1695--1724, Dec.
  2013.

\bibitem{Branke2016}
J.~Branke, S.~Nguyen, C.~W. Pickardt, and M.~Zhang, ``Automated design of
  production scheduling heuristics: A review,'' \emph{IEEE Transactions on
  Evolutionary Computation}, vol.~20, no.~1, pp. 110--124, 2016.

\bibitem{Deb2002}
K.~Deb, A.~Pratap, S.~Agarwal, and T.~Meyarivan, ``A fast and elitist
  multiobjective genetic algorithm: Nsga-ii,'' \emph{IEEE Transactions on
  Evolutionary Computation}, vol.~6, no.~2, pp. 182--197, 2002.

\bibitem{Tricoire2018}
F.~Tricoire, J.~Scagnetti, and A.~Beham, ``{New insights on the block
  relocation problem},'' \emph{Computers and Operations Research}, vol.~89, pp.
  127--139, 2018.

\end{thebibliography}


%

\begin{IEEEbiography}[{\includegraphics[width=1in,height=1.25in,clip,keepaspectratio]{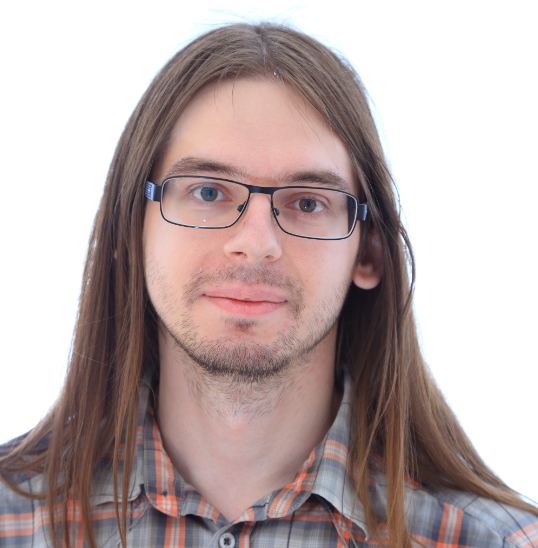}}]{Marko Đurasević}
received his Ph.D. degree in computer science from the University of Zagreb, Faculty of Electrical Engineering and Computing. He is currently an Assistant Professor at the same university. He has published 18 journal and conference papers. His main research interest is the application of hyper-heuristic methods for various scheduling problems. He serves as a reviewer for several international journals and conferences. 
\end{IEEEbiography}

\begin{IEEEbiography}[{\includegraphics[width=1in,height=1.25in,clip,keepaspectratio]{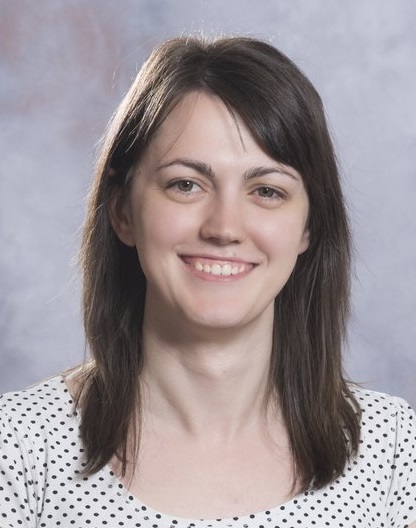}}]{Mateja Đumić}
Mateja Đumić is a Ph.D. researcher at the Department of Mathematics, University of Osijek. She received her Ph.D. degree in May 2020 at the Faculty of Electrical Engineering and Computing, University of Zagreb. Her main research interests are scheduling problems and evolutionary computing. 
\end{IEEEbiography}




\end{document}